\pdfoutput=1
\documentclass[11pt]{article}

\usepackage[]{EMNLP2022}

\usepackage{times}
\usepackage{latexsym}

\usepackage[T1]{fontenc}
\usepackage[utf8]{inputenc}

\usepackage{microtype}

\usepackage{inconsolata}

\usepackage{graphicx}
\usepackage{float}

\title{BERT in Plutarch's Shadows}

\author{Ivan P. Yamshchikov \\
  Max Planck Institute for\\
 Mathematics in the Sciences\\
 Leipzig, Germany\\
 CEMAPRE, \\
 University of Lisbon, Portugal\\
  \texttt{ivan@yamshchikov.info} \\\And
   Alexey Tikhonov\\
Inworld.AI \\
Berlin, Germany \\
\texttt{altsoph@gmail.com} \\\And
Yorgos Pantis\\
Max Planck Institute for\\
Mathematics in the Sciences\\
Leipzig, Germany\\
\AND 
Charlotte Schubert\\
University of Leipzig \\
Leipzig, Germany\\\And
J{\"u}rgen Jost\\
 Max Planck Institute for\\
 Mathematics in the Sciences\\
 Leipzig, Germany\\}
 
\begin{document}
\maketitle
\begin{abstract}
The extensive surviving corpus of the ancient scholar Plutarch of Chaeronea (ca. 45-120 CE) also contains several texts which, according to current scholarly opinion, did not originate with him and are therefore attributed to an anonymous author Pseudo-Plutarch. These include, in particular, the work Placita Philosophorum (Quotations and Opinions of the Ancient Philosophers), which is extremely important for the history of ancient philosophy. Little is known about the identity of that anonymous author and its relation to other authors from the same period. This paper presents a BERT language model for Ancient Greek. The model discovers previously unknown statistical properties relevant to these literary, philosophical, and historical problems and can shed new light on this authorship question. In particular, the Placita Philosophorum, together with one of the other Pseudo-Plutarch texts, shows similarities with the texts written by authors from an Alexandrian context (2nd/3rd century CE).
\end{abstract}

\noindent "I do not need a friend who changes when I change and who nods when I nod; my shadow does that much better." (Plutarch, Quomodo adulator ab amico internoscatur 53b 10)

\section{Introduction}

\noindent Authorship attribution through some form of statistical inference dates at least
half of a century back, when Mosteller and Wallace used statistics of short, frequent words to estimate authorship of Federalist Papers, disputed by \citet{mosteller1963inference}. The physicist \citet{fucks1968} was the first who systematically developed the methods that relied on statistical patterns like word or sentence length that have been found to be able to distinguish between different authors. While undeniably successful in many cases, they do not require any deeper understanding of the texts in question and therefore have their natural limitations. For a detailed review of various authorship attribution methods developed further, we refer the reader to \cite{stamatatos2009survey}. These methods rely on statistical patterns like word or sentence length that have been found to be able to distinguish between different authors. While undeniably successful in many cases, they do not require any deeper understanding of the texts in question and therefore have their natural limitations.

Transformer artificial neural networks have shown spectacular success in machine translation and answering search queries by internally extracting, after extensive pretraining, abstract patterns, and long-range dependencies in human language samples through encoder hierarchies \cite{vaswani2017attention}. Therefore, it seems natural to apply such schemes to other tasks that traditionally depended on human language understanding. Thus, this paper uses versions of BERT to investigate questions of authorship in Ancient Greek literature. Such models have already been used for authorship attribution, but the
actual number of cases is still limited. \citet{fabien2020bertaa} demonstrate that fine-tuning of a pretrained BERT  language model with an additional dense layer and a softmax activation allows performing authorship classification. \citet{polignano2020contextualized} use BERT for author profiling in social media and conclude that despite encouraging results in terms of reliability, the computational power required for running such a model is too demanding for the task. These results show that transformers could be successfully used for authorship attribution, yet the application of these models to historical texts is still limited. Specifically, \citet{bamman2020latin} develop BERT for Latin and \cite{assael2019restoring} train an LSTM language model of Ancient Greek. There is also a char-BERT implementation\footnote{https://github.com/brennannicholson/ancient-greek-char-bert}, but we do not know of any public full BERT model trained to work with Ancient Greek. We are also unaware of any examples when BERT was used successfully for authorship attribution of historical documents. 

For this paper, we focus on Plutarch of Chaeronea (ca. 45-120 CE), a Greek philosopher and biographer. His parallel biographies (one Greek and one Roman) and his philosophical-ethical writings, which he wrote in the tradition of Plato, intending to establish a coherent system, have been widely read in ancient and modern times. This paper addresses the authorship attribution of three manuscripts attributed in antiquity to Plutarch: "De Fluviis", "De Musica", and "Placita Philosophorum" (abstracts and quotations from the now lost works of the ancient philosophers or schools). Classicists argue that these three texts were not written by
Plutarch himself, yet the actual author(s) of these manuscripts is/are not known, and in particular, the question of authorship of the Placita Philosophorum has been widely discussed \cite{mansfeld1997, mansfeld2009a, mansfeld2018, mansfeld2020}. So far, no decisive philological proof could resolve this question, and therefore evidence achieved through modern language processing techniques constitutes a valuable addition to this debate.
 
The contributions of this paper are as follows:
\begin{itemize}
    \item we use a transfer learning approach to train BERT for Ancient Greek;
    \item we demonstrate that this language model is useful for authorship attribution of Ancient Greek texts;
    \item we obtain results that may be used as evidence in the process of authorship attribution of the Pseudo-Plutarchean texts;
    \item we obtain new insights into the paths along which the reception of ancient philosophy was developed.
\end{itemize}

\section{Data}
%%%%%%%%%%%%%%%%%%%%%%%%%%%%%%%%%%%%%%%%%
The data were taken from the digital history projects at the Chair of
Ancient History of the University of Leipzig. The Plutarch texts were transformed into a digital representation under professional supervision. Data can be obtained from the following free repositories Perseus Digital Library\footnote{https://github.com/PerseusDL/canonical-greekLit} and First Thousand Years of Greek\footnote{https://github.com/ThomasK81/TEItoCEX} as part of Open Greek and Latin\footnote{https://opengreekandlatin.org}. The resulting representation was stored in XML format (TEI guidelines) and enriched with metadata. The XML structure and metadata were removed. The strings were transferred into lowercase letters. The diacritics were removed. We did not touch hyphenation, punctuation, or any multilingual remains, nor did we apply any special language-related transformations.
The resulting data set consists of 1 244 documents with 199 809 paragraphs or 14 373 311 words.
%%%%%%%%%%%%%%%%%%%%%%%%%%%%%%%%%%%%%%c%%%

\section{Ancient Greek BERT}

The resulting amount of data was too small to train Ancient Greek BERT from scratch. However, data sets of smaller sizes are routinely used for transfer learning and fine-tuning of transformers. Thus, we suggest obtaining BERT for Ancient Greek via transfer learning on a Masked Language Modelling (MLM) task. One could either use Multilingual BERT\footnote{102 languages including Modern Greek, 110M parameters, see https://github.com/google-research/bert/blob/master/multilingual.md} or Greek BERT\footnote{Modern Greek language, 110M parameters, see
https://huggingface.co/nlpaueb/bert-base-greek-uncased-v1} as a starting model for knowledge transfer. The resulting model could then further be fine-tuned for the task of authorship attribution in Ancient Greek.

\subsection{Tokenizers}

The tokenization of words into sub-word tokens is a crucial preprocessing step that can affect the performance of the model. Up to this point, we were using "words" as a linguistic term, whereas we understand tokens as the output of a tokenization algorithm. Thus, there would be one or more tokens that represent every word. Counting the number of tokens used on average to represent a word or, conversely, an average number of words per token gives an estimate of how fit  the tokenization is for the data set. In particular, the average number of words per token varies from 0 to 1, and the closer it is to 1, the more words are represented with one token. Various researchers have shown that corpus-specific tokenization could be beneficial for an NLP task. For example, \citet{sennrich-etal-2016-neural} show that optimal vocabulary is dependent on the frequencies of the words in the target corpus. \citet{lakew2019controlling} and \citet{aji2020neural} partially discuss the tokenization in the setting of cross-language transfer. Though \citet{aji2020neural} demonstrate that there is no clear evidence that one parent is better than another for cross-lingual transfer learning, they also show that token matching and joint vocabulary \cite{nguyen2017transfer} are the best ways to handle the embedding layer during transfer learning. 

Since each model has its own specific tokenizer before pretraining, one wants to measure how well each of them works with Ancient Greek. To do that, one could take two sample data sets: a sample of Modern Greek
Wikipedia (referred to in Table \ref{tab:tok} as “modern”) and a comparable sample of Ancient Greek (“ancient”  in Table \ref{tab:tok}). Each data sample is tokenized with both the Modern Greek BERT tokenizer\footnote{https://huggingface.co/nlpaueb/bert-base-greek-uncased-v} and the Multilingual BERT tokenizer\footnote{https://huggingface.co/bert-base-multilingual-cased}. One can speculate that the model uses shorter tokens to adopt grammatical information and deal with longer, rarely observed words. In contrast, the representations with longer tokens could be useful for semantically intensive problems. These longer, semantically charged tokens may vary significantly on various downstream tasks. Thus, the average length of a token and the average number of words per token, shown in Table \ref{tab:tok}, could be coarsely used as an estimate of the resulting tokenization. One could claim that the higher these values, the more apt the tokenizer is for the task. Indeed, higher average length of a token and number of words per token mean that longer, more semantically charged tokens could be matched for transfer; see \cite{singh2019bert,aji2020neural,samenko2021fine} for a detailed discussion of various tokenization properties.

\begin{table*}
\centering
\begin{tabular}{lrrrr}
\hline
              & \multicolumn{2}{c}{Symbols per Token}                                  & \multicolumn{2}{c}{Words per Token}                                    \\ \cline{2-5} 
Tokenizer     & \multicolumn{1}{l}{Greek BERT} & \multicolumn{1}{l}{Multilingual BERT} & \multicolumn{1}{l}{Greek BERT} & \multicolumn{1}{l}{Multilingual BERT} \\ \cline{2-5} 
Modern Greek  & 4.52                           & 2.55                                  & 0.72                           & 0.41                                  \\
Ancient Greek & 2.98                           & 1.9                      & 0.46                           & 0.31                                  \\ \hline
\end{tabular}
\caption{In comparison with multilingual BERT, Greek BERT tokenizer shows a  higher number of symbols and words per token for both Modern and Ancient Greek} \label{tab:tok}
\end{table*}

It is hard to compare the resulting tokenizations that we obtain for the same vocabulary size. Some of the tokens occur in both tokenizations, yet have different frequencies, and some are unique for one of the tokenizations. We want to emphasize that direct token matching would not be relevant for comparing the models. Indeed, a frequent token not matching might significantly influence the downstream performance, while several low-frequency non-matching tokens might not have any noticeable effect on the downstream performance. For a detailed comparison, we publish the resulting vocabularies along with relative frequencies of the tokens obtained\footnote{https://clck.ru/32HWhK}. 

Looking at Table \ref{tab:tok}, one could conclude that the tokenizer of Modern Greek BERT is a preferable solution for Ancient Greek. However, the higher number of symbols or words per token does not automatically guarantee that the overall performance of the model
after fine-tuning would be superior in terms of performance on a downstream task.

\subsection{Training Ancient Greek BERT via Transfer Learning}

If related tasks are available, we can fine-tune the model first on a related task with more data before fine-tuning it on the target task, see \cite{ruder2019transfer}. This helps particularly for tasks with limited data \cite{phang2018sentence} and improves sample efficiency on the target task \cite{yogatama2019learning}. Since we have a limited amount of Ancient Greek texts, we want to do language transfer from Modern Greek to Ancient Greek training a masked language
model (MLM)\footnote{https://github.com/huggingface/transformers/blob/master/ examples/language-modeling/run\_language\_modeling.py} of the Ancient Greek text. 

After splitting the original documents in Ancient Greek into semi-sentences with nltk.sent\_tokenize() we obtain 162 490 lines of text for MLM  training. With a learning rate of $1e-4$ and a block size of $512$  we ran MLM on the obtained data set once to avoid any overfitting. As mentioned, the models use different tokenizers, so we cannot compare their performance directly regarding the MLM loss. Since the ultimate goal of this project is authorship attribution, it makes sense to compare the resulting models in terms of the accuracy of the resulting author classifiers that one could build on top of the BERT after MLM transfer learning of Ancient Greek.

\subsection{Authorship Attribution with Ancient Greek BERT}

Let us now check whether BERT after transfer learning via MLM on Ancient Greek texts can be further fine-tuned for authorship attribution. For that purpose, we build an authorship attribution data set using the sixteen most prolific authors of the period in question, the 1st-3rd century CE: Galenus, Origenes, Plutarch, Cassius Dio, Flavius Josephus, Philo Judaeus, Athenaeus, Claudius Ptolemaeus, Aelius Aristides, Strabo, Lucianus, Clemens Alexandrinus, Appianus, Pausanias, Sextus Empiricus, Dio Chrysostomus. Texts by Pseudo-Plutarch are not included in the authorship attribution data set, yet the obtained classifier would be further used to analyze these impersonated documents.

From Dio Chrysostomus we had only 5580 sentences available for training, but this amount of data could be sufficient \cite{zhang2020revisiting} to train a well-performing BERT-based classifier. For the author classifier, to avoid bias from different sentences,  we sample 5 580 sentences from every author in this list. We also include the  label "Others" for 5 580
random sentences by less prolific authors who were not included in this shortlist. The choice of the number of authors balances the requirement to have enough data to train the classifier and the wish to include as many authors in the authorship attribution classifier as possible. 

We thus have seventeen categories (sixteen authors and one extra category denoted as "Others") with 5 580 sentences in each. Five thousand eighty sentences out of every category are used for fine-tuning, while five hundred random sentences in every category are set aside for validation. 

We use this data set to train BERT Classifiers similarly to \cite{fabien2020bertaa}. Table \ref{tab:au} shows the validation accuracy for Modern Greek and Multilingual BERTs after MLM on Ancient Greek and 10 epochs of classifier training\footnote{AdamW, LR = 2e-5, eps = 1e-8,  linear\_schedule}. Table \ref{tab:au}  also shows a standard NLTK Naive Bayes Classifier trained on the 2 000 most frequent unigrams as a reference point for authorship attribution accuracy.

\begin{table}[h]
\centering
\begin{tabular}{lr}
\hline
& Validation\\
& accuracy  \\ 
\hline
Greek BERT  & \textbf{80\%}                                              \\
Greek BERT no MLM-transfer  & 78\%                                              \\
Multilingual  BERT                 & 78\%                                                  \\ 
Naive Bayes Classifier &   43\%                                                 \\ 
Random authorship attribution &   6\%                                                 \\ 
\hline
\end{tabular}
\caption{After MLM training and ten epoch of fine-tuning for authorship attribution, the validation accuracy of  Modern Greek BERT  is slightly higher than that of the Multilingual BERT after similar fine-tuning procedures. Modern Greek BERT fine-tuned for authorship attribution without MLM transfer learning phase shows lower validation accuracy. All BERT-based classifiers significantly outperform the Naive Bayes Classifier that uses the two thousand most frequent unigrams. Another baseline attributes one of seventeen labels to the text at random.}
\label{tab:au}
\end{table}

Though all BERT-based classifiers show comparable validation accuracy, Modern Greek BERT after MLM on ancient texts is slightly better than Multilingual BERT in terms of validation accuracy. Fine-tuning Modern Greek BERT for authorship attribution without MLM transfer learning phase also provides lower validation accuracy in comparison with the combination of MLM transfer and fine-tuning. 

Since Modern Greek BERT fine-tuned for authorship attribution after MLM shows the best validation accuracy, we use it for the subsequent analysis. Table \ref{tab:ac} shows the confusion matrix of the resulting classifier on 500 validation sentences by every author.

\begin{table*}[h]
\centering
{
\small
\tabcolsep=0.11cm
\begin{tabular}{lrrrrrrrrrrrrrrrrr}
                                                               & G            & O            & P            & CD           & FJ           & PJ           & A            & CP           & AA           & S            & L            & CA           & Ap           & P            & SE           & DC           & other        \\ \hline
Galenus                                                        & \textbf{416} & 7            & 5            & 1            & 1            & 4            & 5            & 5            & 6            & 3            & 3            & 5            & 1            & 0            & 6            & 10           & 22           \\ \hline
Origenes                                                       & 2            & \textbf{396} & 0            & 1            & 6            & 4            & 5            & 12           & 1            & 3            & 0            & 24           & 0            & 0            & 6            & 5            & 35           \\ \hline
Plutarchus                                                     & 6            & 3            & \textbf{390} & 3            & 9            & 8            & 17           & 2            & 2            & 5            & 2            & 5            & 12           & 1            & 6            & 13           & 16           \\ \hline
\begin{tabular}[c]{@{}l@{}}Cassius\\ Dio\end{tabular}          & 1            & 0            & 8            & \textbf{428} & 5            & 2            & 2            & 0            & 7            & 8            & 2            & 1            & 17           & 6            & 0            & 7            & 6            \\ \hline
\begin{tabular}[c]{@{}l@{}}Flavius\\ Josephus\end{tabular}     & 3            & 3            & 10           & 5            & \textbf{418} & 2            & 4            & 6            & 8            & 9            & 6            & 1            & 8            & 0            & 4            & 9            & 4            \\ \hline
\begin{tabular}[c]{@{}l@{}}Philo\\ Judaeus\end{tabular}        & 5            & 10           & 13           & 3            & 16           & \textbf{403} & 3            & 3            & 2            & 3            & 3            & 12           & 0            & 0            & 11           & 8            & 5            \\ \hline
Athenaeus                                                      & 11           & 6            & 17           & 4            & 4            & 2            & \textbf{368} & 4            & 7            & 11           & 9            & 7            & 2            & 6            & 6            & 14           & 22           \\ \hline
\begin{tabular}[c]{@{}l@{}}Claudius \\ Ptolemaeus\end{tabular} & 3            & 0            & 0            & 0            & 0            & 1            & 0            & \textbf{480} & 0            & 8            & 0            & 0            & 0            & 0            & 5            & 0            & 3            \\ \hline
\begin{tabular}[c]{@{}l@{}}Aelius\\ Aristides\end{tabular}     & 7            & 6            & 6            & 6            & 7            & 2            & 5            & 0            & \textbf{368} & 8            & 10           & 6            & 1            & 3            & 3            & 40           & 22           \\ \hline
Strabo                                                         & 4            & 5            & 9            & 0            & 3            & 2            & 7            & 1            & 9            & \textbf{432} & 4            & 1            & 3            & 6            & 4            & 4            & 6            \\ \hline
Lucianus                                                       & 2            & 3            & 6            & 1            & 5            & 4            & 9            & 0            & 13           & 9            & \textbf{360} & 12           & 5            & 6            & 6            & 30           & 29           \\ \hline
\begin{tabular}[c]{@{}l@{}}Clemens\\ Alexandrinus\end{tabular} & 8            & 28           & 3            & 4            & 10           & 14           & 4            & 1            & 6            & 5            & 8            & \textbf{349} & 0            & 5            & 17           & 11           & 27           \\ \hline
Appianus                                                       & 1            & 0            & 10           & 18           & 8            & 2            & 2            & 1            & 3            & 2            & 5            & 3            & \textbf{437} & 0            & 0            & 3            & 5            \\ \hline
Pausanias                                                      & 0            & 1            & 1            & 2            & 0            & 0            & 2            & 0            & 4            & 3            & 2            & 3            & 2            & \textbf{472} & 0            & 3            & 5            \\ \hline
\begin{tabular}[c]{@{}l@{}}Sextus\\ Empiricus\end{tabular}     & 2            & 4            & 6            & 0            & 1            & 1            & 4            & 1            & 2            & 2            & 1            & 11           & 0            & 0            & \textbf{446} & 7            & 12           \\ \hline
\begin{tabular}[c]{@{}l@{}}Dio\\ Chrysostomus\end{tabular}     & 2            & 4            & 12           & 9            & 5            & 3            & 7            & 0            & 9            & 10           & 10           & 9            & 6            & 4            & 2            & \textbf{398} & 10           \\ \hline
other                                                          & 17           & 23           & 22           & 7            & 6            & 15           & 32           & 14           & 10           & 6            & 12           & 18           & 6            & 9            & 40           & 21           & \textbf{242} \\ \hline
\end{tabular}}
\caption{The confusion matrix of the obtained authorship classifier. Every horizontal line sums up to 500 sentences by the corresponding author that were set aside for validation. Every column shows the number of sentences labelled by classifier as sentences authored by the corresponding author.}
\label{tab:ac}
\end{table*}

Table \ref{tab:ac} shows that the authorship attribution model works rather well. The errors mostly happen in sentences that have a topical affinity to another author or on authors with similar regional backgrounds. That observation suggests developing a separate regional classifier that might help authorship attribution. This classifier is described in detail in the next Subsection.

\subsection{Regional Attribution with Ancient Greek BERT}

\begin{figure}[h!]
\centering
     \includegraphics[scale=0.35]{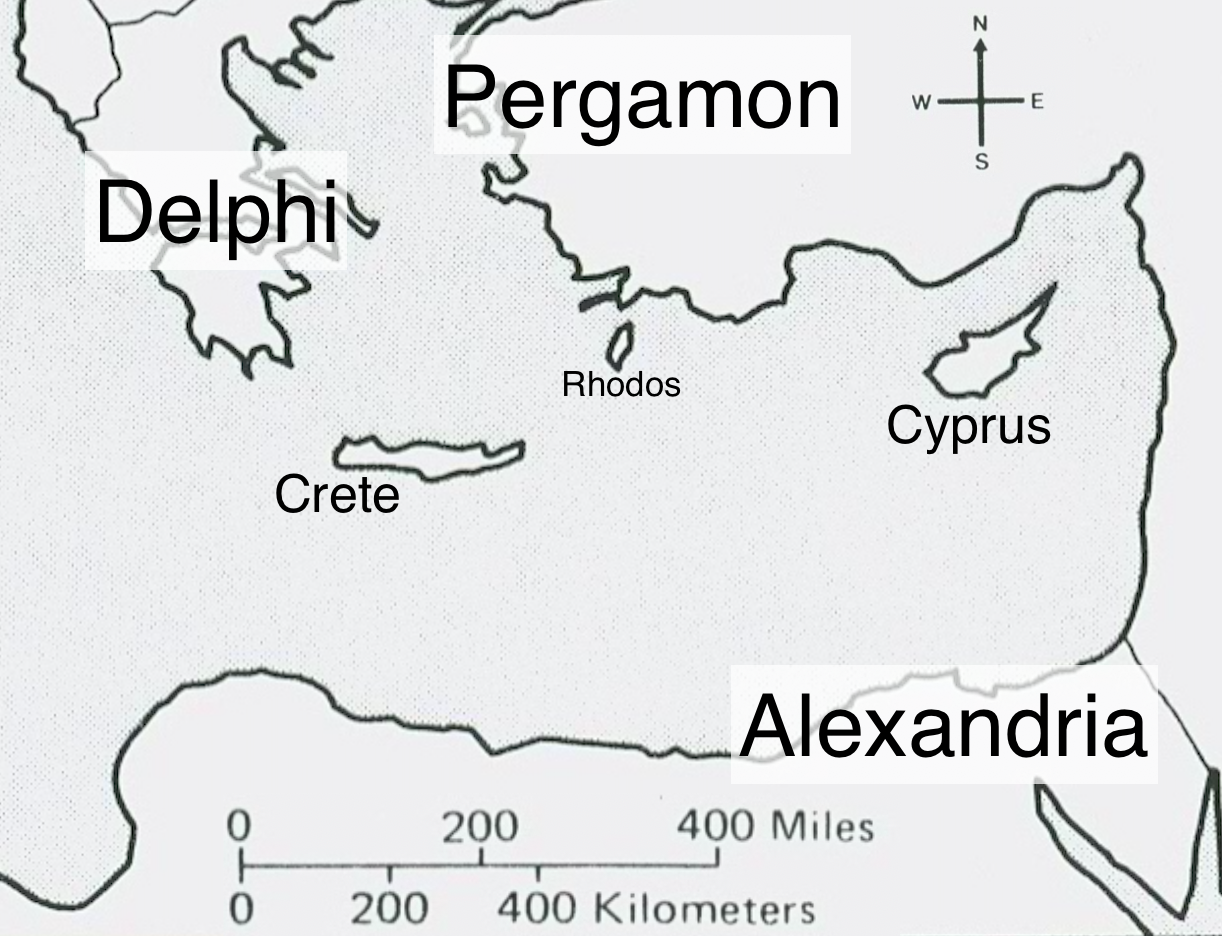}
  \caption{A map showing relative position on three potential regions relevant for authorship attribution of Pseudo-Plutarch documents.}
   \label{fig:map}
\end{figure}

Since Ancient Greek was used in various regions and territories, one might expect that the texts also show regional peculiarities. Such peculiarities
then should also be detectable by a dedicated regional BERT Classifier. We constructed three coarse regions for the origins of the authors in the data set: a region surrounding Delphi, where Plutarch was working, a region in the proximity of Alexandria, and the region of ancient Ionia, namely the ancient region on the central part of the western coast of modern Anatolia, see Figure \ref{fig:map}. After balancing texts written by authors in these three regions with the fourth label that includes random sentences from authors outside of these regions, we train another BERT-based classifier to achieve a $0.79$ validation accuracy for the region of the author. Table \ref{tab:region} shows the results of the obtained classifier on the validation set.

\begin{table*}[h]
\centering
\begin{tabular}{lrrrr}
\hline
               \multicolumn{1}{c}{Predicted Region}      & \multicolumn{1}{c}{Pergamon Region} & \multicolumn{1}{c}{Alexandria Region} & \multicolumn{1}{c}{Delphi Region} & \multicolumn{1}{c}{Other Regions} \\ \hline
 Pergamon   & \textbf{83\%}                               & 3\%                                  & 3\%                             & 7\%                              \\
 Alexandria & 5\%                                & \textbf{77\%}                                 & 7 \%                             & 10\%                               \\
 Delphi      & 4\%                                & 5\%                                   & \textbf{81\%}                             & 8\%                              \\
 Other      & 8\%                                & 15\%                         & 9\%                             & \textbf{75\%}                             \\ \hline
\end{tabular}
\caption{Results of the BERT-based regional classifier on 4000 sentences set aside for validation.}
\label{tab:region}
\end{table*}

With the author classifier having $80\%$ validation accuracy on eighteen author categories and the regional classifier having $79\%$ validation accuracy on four regional categories, we can try to get some insights into the origins of the Pseudo-Plutarch texts.

\section{Classifying Pseudo-Plutarch}

Our stated aim was to obtain further insights into the authorship attribution of three manuscripts attributed to Plutarch in antiquity: "De Fluviis", "De Musica", and "Placita Philosophorum". Though classicists argue that these three texts were not written by Plutarch himself, the actual author(s) of these manuscripts is/are not known. Thus any new insights into the authorship of these documents might be useful to advance classical philology and ancient history.

Let us now split these three Pseudo-Plutarchean texts into the separate sentences and apply the author classifier described above to these texts.
Table \ref{tab:p} shows the authors that are most frequently attributed within a particular document, along with the share of sentences attributed to them. We have double-checked these results using an alternative scoring method. Instead of classifying every sentence in a document and then averaging the classifier's results across all sentences, one could obtain probability scores for every author that the model estimates for every sentence. Averaging those probabilities throughout the document, one could obtain the three most-probable author candidates. These three most-probable authors turn out to be exactly the same for all three documents as the ones in Table \ref{tab:p}. Moreover, the resulting probabilities of the authorship are also the same for all three most probable authors across all three documents under examination.

\begin{table*}[]
\centering
\begin{tabular}{lrrrrrrr}
\hline
		& \multicolumn{1}{l}{Sample} 	& \multicolumn{1}{l}{Top 1} 	& \multicolumn{1}{l}{Top 1} 	& \multicolumn{1}{l}{Top 2} 	& \multicolumn{1}{l}{Top 2} 	& \multicolumn{1}{l}{Top 3} 	& \multicolumn{1}{l}{Top 3} \\ 
		& \multicolumn{1}{l}{Size} 	& 	& \multicolumn{1}{l}{Share} 	& 	& \multicolumn{1}{l}{Share} 	& & \multicolumn{1}{l}{Share} \\ 

\hline
De Fluviis  & 310 						& Athenaeus 				& 22\% 						& Others 					& 21\% 						& Strabo 					&  19\%                                       \\
\hline
   & 						&  				&  						&				& 					& Sextus  			&             \\ 
De Musica   & 285 						& Athenaeus 				& 21\% 						& Plutarch 				& 18\%						& Empiricus 			&     14\%         \\ 
\hline
Placita  &  				&  					&						& Claudius  		&  						& Sextus  			&    \\
Philosophorum  & 928 				& Others 					& 36\% 						& Ptolemaeus  		& 20\% 						& Empiricus  			& 11 \%   \\
\hline
\end{tabular}
\caption{The most frequently attributed authors in the three Pseudo-Plutarchean texts.}
\label{tab:p}
\end{table*}

Table \ref{tab:p}  demonstrates that the resulting authorship profiles differ for all three documents. However, the sample size for De Fluviis and De Musica is smaller than the sample size for Placita Philosophorum. Plutarch ends up being in the top three of the most frequently predicted authors only in De Musica. For De Fluviis and De Musica Athenaeus ends up being the most frequent guess of the BERT authorship classifier, while for Placita Philosophorum "other" is the most frequent attribution. Claudius Ptolemaeus and Sextus Empiricus are, respectively, the second and the third most frequent guesses. 

\begin{figure}[h!]
\centering
     \includegraphics[scale=0.55]{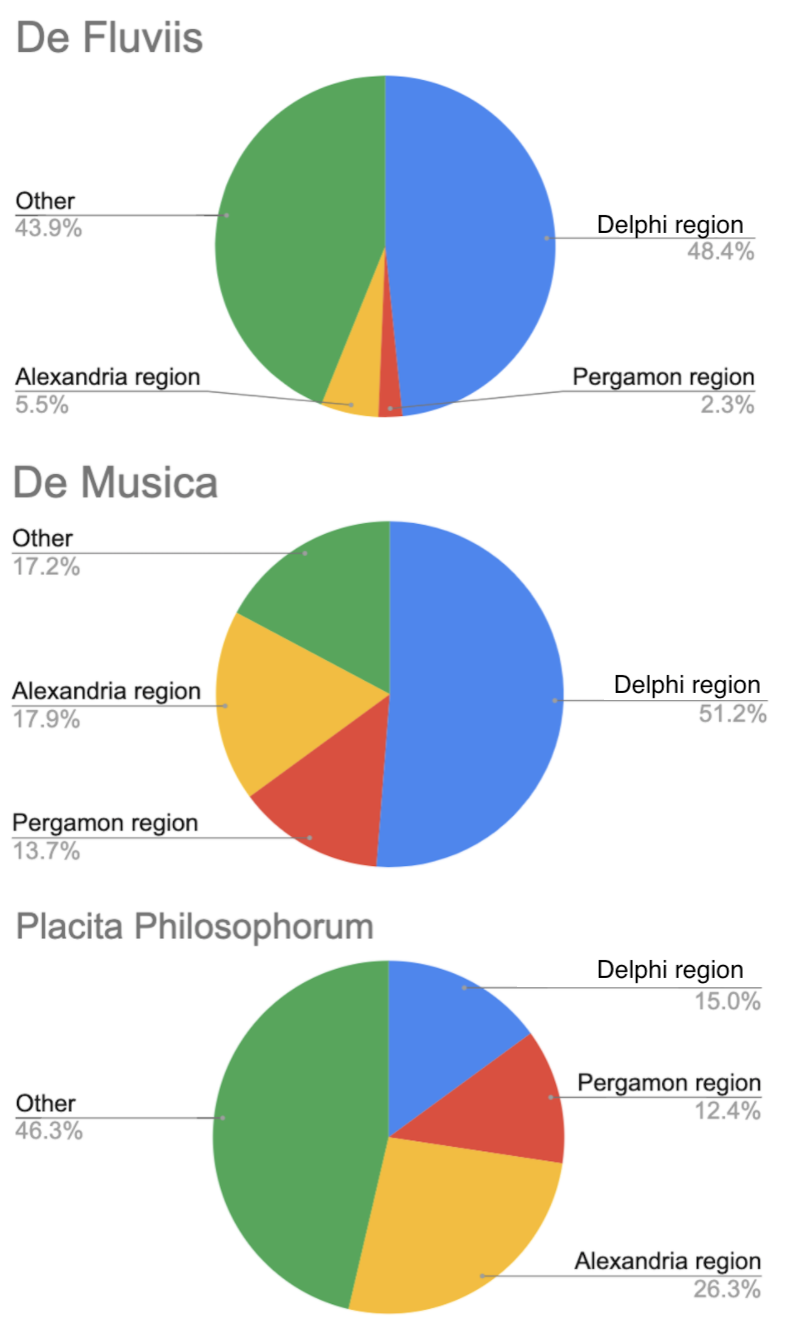}
  \caption{All three Pseudo-Plurach documents show significantly different percentages of sentences attributed to a certain region. In particular, Placita Philosophorum is the only document where Delphi is not a dominant region, while Alexandria is the most frequent identifiable region.}
   \label{fig:reg}
\end{figure}

Figure \ref{fig:reg} shows the regional profile obtained with the BERT-based regional classifier. Once again, all three works show different structural properties. While the model associates every second sentence from De Fluviis and De Musica with the Delphi region, it only attributes $15\%$ of the sentences from Placita Philosophorum to Delphi. The model is far more uncertain about the third document. Every fourth sentence in Placita Philosophorum is associated with the Alexandrian region, while almost half is labeled as "other".

All in all, the result shows that the Placita Philosophorum is closely related to a philosophical-scientific tradition from the 1st century CE to ca. 220/250 CE. Possibly one could even see an embedding into a specifically Alexandrian context, which despite very different contents of the works (Strabo as a geographer, Ptolemy as mathematician and geographer, Athenaeus as an anthologist, and Sextus as a Skeptic), is related to the Pseudo-Plutarchian Placita. The parallels between Pseudo-Plutarch and Sextus Empiricus have already been pointed out several times in the literature \cite{mansfeld2020}; connections to Claudius Ptolemaeus have not been considered so far. It is fascinating that in this group of three (Pseudo-Plutarch, Claudius Ptolemaeus, and Sextus
Empiricus), a mathematical-cosmologically oriented context for very different topics of philosophy, natural science, and medicine, is the basis. We can thus state a new hypothesis that the Placita Philosophorum presumably originates from an Alexandrian scientific context of the 2nd century CE. We also should notice that De Fluviis and De Musica, and Placita Philosophorum might have three different authors. There are some similarities between De Fluviis and De Musica (both have approximately 50\% of the text attributed to the Delphi region and approximately 20\% attributed to Athenaeus). At the same time, one should be cautious in the conclusions since Placita Philosophorum is longer and have a different topical structure than the other two works. This might explain the lower percentage of Delphi-attributed sentences in the document. Athenaeus was allegedly born on the territory of modern Egypt, as well as Claudius Ptolemaeus.

Sextus Empiricus lived in different places, however, like the authors of De Fluviis and
De Musica, he seems to have been more influenced by the Alexandrian tradition than has been seen so far. Furthermore, the results  shown in Table  \ref{tab:au}, and in figure \ref{fig:reg}  indicate that the three works analyzed here (De Fluviis, De Musica, Placita Philosophorum) cannot be securely linked to one identifiable author. Since all three works have a strongly compilatory character due to the many  quotations, references and summaries, this result is quite plausible and confirms the current scholarly opinion against an attribution to Plutarch.

We hope that this work may inspire or guide further philological and historical
research on this problem. Authorship attribution helps to contextualize certain
historical and philosophical ideas and better understand their development and
influence on each other. This is especially important for the Placita Philosophorum, as
they are one of the most important sources for ancient philosophy, medicine and
cosmology.

\section{Discussion}

\citet{koppel2009computational} classify fundamental problems that arise when researchers try to establish authorship via statistical inference. One of the
fallacies associated with such research is the so-called "needle in a haystack" fallacy. This may arise when the number of potential author candidates is exceedingly large, yet these authors are not necessarily represented in the training data. We are aware of this problem converting authorship attribution of Pseudo-Plutarchean texts with regard to the complicated and often fragmentary transmission of ancient texts as well as the complex research discussion \cite{mansfeld1997,mansfeld2009a,mansfeld2009b,mansfeld2018,mansfeld2020}. However, this paper provides new meaningful insights into the possible relationships
and background of these texts. In particular, we have obtained evidence that the pseudo-Plutarchian texts investigated here did not originate from  Plutarch himself, and we could narrow down the intellectual context, both concerning the time and the region, from which they most likely arose, although none of the other authors that we have used for comparison emerges as Pseudo-Plutarch. We have also detected systematic relations between authors from the 1st-3rd century CE that should merit closer philological analysis. %These results have also been corroborated with traditional statistical tools (StyloAH and Gephi; results not shown here).
  
We publish the weights of the obtained BERT for Ancient Greek\footnote{https://huggingface.co/altsoph/bert-base-ancientgreek-uncased} and hope it will facilitate further applications of modern language models to ancient texts. We are also working on a detailed follow-up that analyzes results obtained using classicist expertise. One has to remember that ancient linguists typically are facing the so-called problem of "small data". Dead languages can have limited data sets available for research. The paper demonstrates that sometimes transfer learning could be a feasible workaround.

Since we are interested in a particular historical time-span around Plutarch, we did not provide details on other potential applications for the resulting model. However, one could list several further applications that might interest historians and, to our knowledge, are not developed to this day. Stylistic attributes of text include author-specific attributes (see \cite{xu} or \cite{Jhamtani} on 'shakespearization'), politeness \cite{Sennrich}, gender or political slant \cite{Prabhumoye}, formality of speech \cite{Rao} but most importantly for the scope of this work the \textit{'style of the time'} \cite{Hughes}. Using the provided approach for the historical dating of the documents is a feasible option that might bring new historical insights. It is only the question of data and their quality. For example, one could try to use a corpus spanning several hundreds of years, fine-tune the developed model for the task of date attribution and see if it could work reasonably well. This is a possible further line of work to pursue, yet it is outside of the scope of this contribution.

\section{Conclusion}

This paper develops BERT for Ancient Greek. It demonstrates that Modern Greek BERT after transfer learning via MLM on Ancient Greek texts could be further fine-tuned as an authorship attribution classifier for ancient texts. The validation accuracy of authorship attribution is shown to be 80\%. The model is then used to analyze text attributed to Pseudo-Plutarch. It shows that three documents have distinctly different statistical properties, and while De Musica and De Fluviis might originate with the same author, Placita Philosophorum has a different authorship and regional profile. Thus, the classification of authorship allows the search for the authors of the 3 works gathered under Pseudo-Plutarch to be narrowed down to the 1st-3rd century CE, suggesting that at least one of the authors may have come from the vicinity of Alexandria

\section*{Limitations}

The research relies on the pre-trained mBERT model as well as the Greek BERT. Yet we believe the proposed fine-tuning procedure could be applicable to other low-resource languages.

GPU is preferable to achieve results within a reasonable time.

\section*{Ethics Statement}
This paper complies with the \href{https://www.aclweb.org/portal/content/acl-code-ethics}{ACL Ethics Policy}.

\section{Acknowledgments}
Ivan Yamshchikov obtained some of the results while working at LEYA Laboratory by Yandex and Higher School of Economics in St. Petersburg. His work was supported by the grant for research centers in the field of AI provided by the Analytical Center for the Government of the Russian Federation (ACRF) in accordance with the agreement on the provision of subsidies (identifier of the agreement 000000D730321P5Q0002) and the agreement with HSE University  No. 70-2021-00139.

\bibliography{plutarch}

\begin{thebibliography}{30}
\expandafter\ifx\csname natexlab\endcsname\relax\def\natexlab#1{#1}\fi

\bibitem[{Aji et~al.(2020)Aji, Bogoychev, Heafield, and
  Sennrich}]{aji2020neural}
Alham~Fikri Aji, Nikolay Bogoychev, Kenneth Heafield, and Rico Sennrich. 2020.
\newblock In neural machine translation, what does transfer learning transfer?
\newblock In \emph{Proceedings of the 58th Annual Meeting of the Association
  for Computational Linguistics}, pages 7701--7710.

\bibitem[{Assael et~al.(2019)Assael, Sommerschield, and
  Prag}]{assael2019restoring}
Yannis Assael, Thea Sommerschield, and Jonathan Prag. 2019.
\newblock Restoring ancient text using deep learning: a case study on greek
  epigraphy.
\newblock In \emph{Proceedings of the 2019 Conference on Empirical Methods in
  Natural Language Processing and the 9th International Joint Conference on
  Natural Language Processing (EMNLP-IJCNLP)}, pages 6368--6375.

\bibitem[{Bamman and Burns(2020)}]{bamman2020latin}
David Bamman and Patrick~J Burns. 2020.
\newblock Latin bert: A contextual language model for classical philology.
\newblock \emph{arXiv preprint arXiv:2009.10053}.

\bibitem[{Fabien et~al.(2020)Fabien, {\'u}~Villatoro-Tello, Motlicek, and
  Parida}]{fabien2020bertaa}
Ma{\"e}l Fabien, Esa {\'u}~Villatoro-Tello, Petr Motlicek, and Shantipriya
  Parida. 2020.
\newblock Bertaa: Bert fine-tuning for authorship attribution.
\newblock In \emph{Proceedings of the 17th International Conference on Natural
  Language Processing}. ACL.

\bibitem[{Fucks(1968)}]{fucks1968}
Wilhelm Fucks. 1968.
\newblock \emph{Nach allen Regeln der Kunst}.
\newblock Deutsche Verlagsanstalt.

\bibitem[{Hughes et~al.(2012)Hughes, Foti, Krakauer, and Rockmore}]{Hughes}
James~M. Hughes, Nicholas~J. Foti, David~C. Krakauer, and Daniel~N. Rockmore.
  2012.
\newblock \href {http://www.pnas.org/content/109/20/7682.full} {Quantitative
  patterns of stylistic influence in the evolution of literature}.
\newblock \emph{Proceedings of the National Academy of Sciences},
  109(20):7682--7686.

\bibitem[{Jhamtani et~al.(2017)Jhamtani, Gangal, Hovy, and Nyberg}]{Jhamtani}
Harsh Jhamtani, Varun Gangal, Eduard Hovy, and Eric Nyberg. 2017.
\newblock \href {http://www.aclweb.org/anthology/W17-4902} {Shakespearizing
  modern language using copy-enriched sequence-to-sequence models}.
\newblock In \emph{Proceedings of the Workshop on Stylistic Variation}, pages
  10--19.

\bibitem[{Koppel et~al.(2009)Koppel, Schler, and
  Argamon}]{koppel2009computational}
Moshe Koppel, Jonathan Schler, and Shlomo Argamon. 2009.
\newblock Computational methods in authorship attribution.
\newblock \emph{Journal of the American Society for information Science and
  Technology}, 60(1):9--26.

\bibitem[{Lakew et~al.(2019)Lakew, Di~Gangi, and
  Federico}]{lakew2019controlling}
Surafel~Melaku Lakew, Mattia Di~Gangi, and Marcello Federico. 2019.
\newblock Controlling the output length of neural machine translation.
\newblock \emph{arXiv preprint arXiv:1910.10408}.

\bibitem[{Mansfeld and Runia(1997)}]{mansfeld1997}
Jaap Mansfeld and David Runia. 1997.
\newblock \emph{A\"etiana: The Method and Intellectual Context of a
  Doxographer, Volume I: The Sources}.
\newblock Philosophia Antiqua 73. Leiden: E.J.Brill.

\bibitem[{Mansfeld and Runia(2009{\natexlab{a}})}]{mansfeld2009a}
Jaap Mansfeld and David Runia. 2009{\natexlab{a}}.
\newblock \emph{A\"etiana: The Method and Intellectual Context of a
  Doxographer, Volume II: The Compendium}.
\newblock Philosophia Antiqua 114. Leiden: E.J.Brill.

\bibitem[{Mansfeld and Runia(2009{\natexlab{b}})}]{mansfeld2009b}
Jaap Mansfeld and David Runia. 2009{\natexlab{b}}.
\newblock \emph{A\"etiana: The Method and Intellectual Context of a
  Doxographer, Volume III: Studies in the Doxographical Traditions of Greek
  Philosophy}.
\newblock Philosophia Antiqua 118. Leiden: E.J.Brill.

\bibitem[{Mansfeld and Runia(2018)}]{mansfeld2018}
Jaap Mansfeld and David Runia. 2018.
\newblock \emph{A\"etiana: The Method and Intellectual Context of a
  Doxographer, Volume IV: Towards an Edition of the Aëtian Placita: Papers of
  the Melbourne Colloquium on Ancient Doxography}.
\newblock Philosophia Antiqua 148. Leiden: E.J.Brill.

\bibitem[{Mansfeld and Runia(2020)}]{mansfeld2020}
Jaap Mansfeld and David Runia. 2020.
\newblock \emph{A\"etiana: Volume V.1–4: An Edition of the text of the
  Placita with a Commentary and a Collection of Related Texts}.
\newblock Philosophia Antiqua 153.1-4. Leiden: E.J.Brill.

\bibitem[{Mosteller and Wallace(1963)}]{mosteller1963inference}
Frederick Mosteller and David~L Wallace. 1963.
\newblock Inference in an authorship problem: A comparative study of
  discrimination methods applied to the authorship of the disputed federalist
  papers.
\newblock \emph{Journal of the American Statistical Association},
  58(302):275--309.

\bibitem[{Nguyen and Chiang(2017)}]{nguyen2017transfer}
Toan~Q Nguyen and David Chiang. 2017.
\newblock Transfer learning across low-resource, related languages for neural
  machine translation.
\newblock In \emph{Proceedings of the Eighth International Joint Conference on
  Natural Language Processing (Volume 2: Short Papers)}, pages 296--301.

\bibitem[{Phang et~al.(2018)Phang, F{\'e}vry, and Bowman}]{phang2018sentence}
Jason Phang, Thibault F{\'e}vry, and Samuel~R Bowman. 2018.
\newblock Sentence encoders on stilts: Supplementary training on intermediate
  labeled-data tasks.
\newblock \emph{arXiv preprint arXiv:1811.01088}.

\bibitem[{Polignano et~al.(2020)Polignano, de~Gemmis, and
  Semeraro}]{polignano2020contextualized}
Marco Polignano, Marco de~Gemmis, and Giovanni Semeraro. 2020.
\newblock Contextualized bert sentence embeddings for author profiling: The
  cost of performances.
\newblock In \emph{International Conference on Computational Science and Its
  Applications}, pages 135--149. Springer.

\bibitem[{Prabhumoye et~al.(2018)Prabhumoye, Tsvetkov, Black, and
  Salakhutdinov}]{Prabhumoye}
Shrimai Prabhumoye, Yulia Tsvetkov, Alan~W. Black, and Ruslan Salakhutdinov.
  2018.
\newblock \href {https://arxiv.org/pdf/1809.06284.pdf} {Style transfer through
  back-translation}.
\newblock In \emph{arXiv preprint}.

\bibitem[{Rao and Tetreault(2018)}]{Rao}
Sudha Rao and Joel Tetreault. 2018.
\newblock \href {http://www.aclweb.org/anthology/N18-1012} {Dear sir or madam,
  may i introduce the gyafc dataset: Corpus, benchmarks and metrics for
  formality style transfer}.
\newblock In \emph{Proceedings of the Conference of the North American Chapter
  of the Association for Computational Linguistics: Human Language
  Technologies}, volume~1, pages 129--140.

\bibitem[{Ruder et~al.(2019)Ruder, Peters, Swayamdipta, and
  Wolf}]{ruder2019transfer}
Sebastian Ruder, Matthew~E Peters, Swabha Swayamdipta, and Thomas Wolf. 2019.
\newblock Transfer learning in natural language processing.
\newblock In \emph{Proceedings of the 2019 Conference of the North American
  Chapter of the Association for Computational Linguistics: Tutorials}, pages
  15--18.

\bibitem[{Samenko et~al.(2021)Samenko, Tikhonov, Kozlovskii, and
  Yamshchikov}]{samenko2021fine}
Igor Samenko, Alexey Tikhonov, Borislav Kozlovskii, and Ivan~P Yamshchikov.
  2021.
\newblock Fine-tuning transformers: Vocabulary transfer.
\newblock \emph{arXiv preprint arXiv:2112.14569}.

\bibitem[{Sennrich et~al.(2016{\natexlab{a}})Sennrich, Haddow, and
  Birch}]{Sennrich}
Rico Sennrich, Barry Haddow, and Alexandra Birch. 2016{\natexlab{a}}.
\newblock \href {http://www.aclweb.org/anthology/N16-1005} {Controlling
  politeness in neural machine translation via side constraints}.
\newblock \emph{In Proceedings of the 2016 Conference of the North American
  Chapter of the Association for Computational Linguistics: Human Language
  Technologies}, pages 35--40.

\bibitem[{Sennrich et~al.(2016{\natexlab{b}})Sennrich, Haddow, and
  Birch}]{sennrich-etal-2016-neural}
Rico Sennrich, Barry Haddow, and Alexandra Birch. 2016{\natexlab{b}}.
\newblock \href {https://doi.org/10.18653/v1/P16-1162} {Neural machine
  translation of rare words with subword units}.
\newblock In \emph{Proceedings of the 54th Annual Meeting of the Association
  for Computational Linguistics (Volume 1: Long Papers)}, pages 1715--1725,
  Berlin, Germany. Association for Computational Linguistics.

\bibitem[{Singh et~al.(2019)Singh, McCann, Socher, and Xiong}]{singh2019bert}
Jasdeep Singh, Bryan McCann, Richard Socher, and Caiming Xiong. 2019.
\newblock Bert is not an interlingua and the bias of tokenization.
\newblock In \emph{Proceedings of the 2nd Workshop on Deep Learning Approaches
  for Low-Resource NLP (DeepLo 2019)}, pages 47--55.

\bibitem[{Stamatatos(2009)}]{stamatatos2009survey}
Efstathios Stamatatos. 2009.
\newblock A survey of modern authorship attribution methods.
\newblock \emph{Journal of the American Society for information Science and
  Technology}, 60(3):538--556.

\bibitem[{Vaswani et~al.(2017)Vaswani, Shazeer, Parmar, Uszkoreit, Jones,
  Gomez, Kaiser, and Polosukhin}]{vaswani2017attention}
Ashish Vaswani, Noam Shazeer, Niki Parmar, Jakob Uszkoreit, Llion Jones,
  Aidan~N Gomez, {\L}ukasz Kaiser, and Illia Polosukhin. 2017.
\newblock Attention is all you need.
\newblock In \emph{Advances in neural information processing systems}, pages
  5998--6008.

\bibitem[{Xu et~al.(2012)Xu, Ritter, Dolan, Grishman, and Cherry}]{xu}
Wei Xu, Alan Ritter, William~B. Dolan, Ralph Grishman, and Colin Cherry. 2012.
\newblock \href {http://www.aclweb.org/anthology/C12-1177} {Paraphrasing for
  style}.
\newblock \emph{Proceedings of COLING}, pages 2899--2914.

\bibitem[{Yogatama et~al.(2019)Yogatama, d'Autume, Connor, Kocisky,
  Chrzanowski, Kong, Lazaridou, Ling, Yu, Dyer et~al.}]{yogatama2019learning}
Dani Yogatama, Cyprien de~Masson d'Autume, Jerome Connor, Tomas Kocisky, Mike
  Chrzanowski, Lingpeng Kong, Angeliki Lazaridou, Wang Ling, Lei Yu, Chris
  Dyer, et~al. 2019.
\newblock Learning and evaluating general linguistic intelligence.
\newblock \emph{arXiv preprint arXiv:1901.11373}.

\bibitem[{Zhang et~al.(2020)Zhang, Wu, Katiyar, Weinberger, and
  Artzi}]{zhang2020revisiting}
Tianyi Zhang, Felix Wu, Arzoo Katiyar, Kilian~Q Weinberger, and Yoav Artzi.
  2020.
\newblock Revisiting few-sample bert fine-tuning.
\newblock In \emph{International Conference on Learning Representations}.

\end{thebibliography}
\bibliographystyle{acl_natbib}

\appendix

\section{Appendix}
\label{appendix}

We have studied two different ways of data splitting to address possible questions on the validity of the proposed machine learning pipeline. The first preprocessing described above splits the texts into sentences and then randomly splits them into training, validation, and testing. This splitting might create some dependencies in the evaluation sets since sentences from the same text could be in training, validation, and testing sets. This potentially could lead to higher evaluation scores, so here we briefly discuss another way of splitting data that could not be prone to such data' leak'. Let us refine our data set leaving only the authors that have produced several documents. Let's do the test-train split so that whole documents end up in the train or the test set. This split limits the list of authors even further, yet it is not a problem in this context. Table \ref{appd:allmodels} summarizes the result on the test set.

\begin{table}[h!]
\centering
\begin{tabular}{lll}
\hline
                             & Validation Acc. \\ \hline
\textbf{Greek BERT with MLM}    &\textbf{83.8\%}  \\
Greek BERT without MLM & 83.4\%  \\
\textbf{M-BERT with MLM}       & \textbf{83.8\%}  \\
M-BERT without MLM     & 81.6\%  \\
Naive Bayes            & 75.8\%   \\
Random Author Assignment  & 11.2\%   \\\hline
\end{tabular}
\caption{Accuracy of the authorship classifier on the test set for the document-balanced splitting.}
\label{appd:allmodels}
\end{table}

Once again, the models fine-tuned with MLM and then trained on the downstream classification show the highest accuracy. The accuracy is even higher than with the document-agnostic sentence-based train-test split. This further illustrates that the sentence-base split is a valid method to train the classifiers.

\end{document}